\documentclass{article}

\usepackage[final]{nips_2018}

\usepackage[utf8]{inputenc} 
\usepackage[T1]{fontenc}    
\usepackage{hyperref}       
\usepackage{url}            
\usepackage{booktabs}       
\usepackage{amsfonts}       
\usepackage{nicefrac}       
\usepackage{microtype}      
\usepackage{geometry}
\usepackage{graphicx}
\usepackage{amsmath}
\usepackage{subfig}

\title{Detecting Offensive Content in Open-domain Conversations using Two Stage Semi-supervision}

\author{
  Chandra Khatri, Behnam Hedayatnia, Rahul Goel\\
  \textbf{Anushree Venkatesh, Raefer Gabriel, Arindam Mandal}\\
  \texttt{chagripk@gmail.com}, \\
  \texttt{\{behnam,goerahul,anuvenk,raeferg,arindamm\}@amazon.com} \\ 
}

\begin{document}

\maketitle

\begin{abstract}
As open-ended human-chatbot interaction becomes commonplace, sensitive content
detection gains importance.
In this work, we propose a two stage semi-supervised approach to bootstrap
large-scale data for automatic sensitive language detection from publicly
available web resources.  We explore various data selection methods including 1)
using a blacklist to rank online discussion forums by the level of their
sensitiveness followed by randomly sampling utterances and 2) training a weakly
supervised model in conjunction with the blacklist for scoring sentences from
online discussion forums to curate a dataset.
Our data collection strategy is flexible and allows the models to detect
implicit sensitive content for which manual annotations may be difficult. We
train models using publicly available annotated datasets as well as using the
proposed large-scale semi-supervised datasets. We evaluate the performance of
all the models on Twitter and Toxic Wikipedia comments testsets as well as on a
manually annotated spoken language dataset collected during a large scale
chatbot competition.
Results show that a model trained on this collected data outperforms the baseline
models by a large margin on both in-domain and out-of-domain testsets, achieving
an F1 score of 95.5\% on an out-of-domain testset compared to a score of 75\%
for models trained on public datasets. We also showcase that large scale two
stage semi-supervision generalizes well across multiple classes of sensitivities
such as hate speech, racism, sexual and pornographic content, etc. without even
providing explicit labels for these classes, leading to an average recall of
95.5\% versus the models trained using annotated public datasets which achieve
an average recall of 73.2\% across seven sensitive classes on out-of-domain
testsets.
 
\end{abstract}

\section{Introduction}
Abusive language is very common on the Internet. Due to the pervasiveness of such language online, sensitive content detection has been an integral problem to solve for social media sites such as Facebook and Twitter. Recently, commercial AI systems such as Amazon's Alexa, Apple's Siri and Google's Assistant are becoming more conversational and respond to a large variety of user utterances. As we move towards knowledge grounded social bots~\citep{hori2018overview}, the monitoring of the knowledge bases that these systems consume also becomes important. 
As these systems are getting ubiquitous, the problem of sensitive content moderation becomes more central to create a better user experience.

Sensitive language is hard to define. On the surface it would seem that a black list of profanities will address the issue but the problem is much broader due to various reasons:
\begin{itemize}
\item Coverage: A blacklist might have high precision but will have a low recall
  as it will miss many abusive comments not present in the list.
\item Cultural differences: A certain word might be fine if used by a particular
  subgroup but might not be acceptable across other subgroups.
\item Context: Personal attacks when viewed as a single utterance might seem
  perfectly innocuous but when viewed within a wider context they become
  sensitive material.
\item Sarcasm: Sarcasm is hard to detect and can hide abuse. Although there has
  been recent work~\citep{oraby2017creating} in detecting sarcasm much progress
  needs to be made before we can detect sarcasm reliably.
\item Expressive Lengthening and non-standard vocabulary: Its easy to get around
  black word filters by changing how the word is spelled while keeping the
  meaning, for example ``f******kkk'' and ``F!ck''.
\item Subjectivity: Sentences like, ``X is the
  capital of Y'' or ``Z is the biggest lie ever told'' could be sensitive for some audiences.
\item Recency: Due to recent political or social events, some things which were
  not sensitive before could become sensitive 
\item Fake News: Fake news has become more ubiquitous in the past few years which can cause confusion around certain content. There has been work in trying to detect fake news~\citep{fakenews}.
\end{itemize}

The problem is exacerbated due to the lack of supervised datasets to train machine learning models to detect various kinds of abuses. The Tay bot~\citep{microsoftTay} provides an important example of why content moderation is necessary. Due to the challenges in solving this issue and the severity of it, companies have typically resorted to humans acting as moderators. However this approach does not scale. Recently,~\cite{curry2018metoo} published a report on how various conversational AI systems respond to sexual harassment, identifying the need for better content moderation on these platforms. Google in partnership with Wikipedia launched the Toxic comment challenge~\citep{toxic} to tackle this problem. 

In this work we propose a bootstrap method to generate a large corpus to train a classification system to detect sensitive content. Our large corpus comes from Reddit\footnote{We use a publicly available Reddit data dump~\citep{reddit}.}which is composed of several subcommunities called subreddits which are topic based. These subreddits are fairly active and the discussions consist of recent sensitive topics like the Middle East which allows our models to combat the recency problem in the sensitive topic detection landscape. 
We take advantage of this natural clustering of communities and adopt a two stage semi-supervised approach, where we start with a seed blacklist of words. We first use the blacklist to rank communities on their levels of sensitiveness. Then we sample from the most sensitive and least sensitive communities to obtain data to train our first classifier. For a second round of better data selection, we leverage the small annotated dataset from the Toxic comment challenge to train a supervised model. This supervised model in conjunction with our blacklist is used again to sample data from our ranked subreddit communities. 

To test the efficiency of our data bootstrap techniques, we test our models on 3 datasets, 2 are public,~\citep{twitter, toxic}, as well as one out-of-domain (open-domain spoken chatbot dataset~\citep{ram2017alexaprize}). Previous work have mostly focused on ``hate speech'' or ``offensive'' language detection. We think ``Sensitive'' language is much broader due to the aforementioned issues and encompasses both hate and offensive speech. We will use the term sensitive speech throughout the rest of our work interchangeably with offensive and hate speech. The datasets we work with are multi-class with categories like hate and offensive for Twitter; toxic, obscene, threat etc for ``Toxic Comments'' and racism, sexual, violence, insult etc for the chatbot dataset. We show that adding the bootstrap data significantly improves performance for in-domain data and generalizes to the out-of-domain data. Furthermore, we provide qualitative examples of the subtle nuances of sensitive content which our models are able to detect. To the best of our knowledge this is the first work in applying sensitive content detection on human-chatbot spoken language data as a use case.

\section{Related Work}\label{related}
The task of identifying sensitive content in natural language has been studied for a long time. Most of the work, unsurprisingly, has focused on identifying hate speech and abusive content on online forums.  
In their seminal work~\cite{spertus1997smokey} built the 'Smokey' system which was a decision tree based on hand engineered semantic and syntactic features for classifying messages as inflammatory or non-inflammatory. One of the earliest works~\citep{yin2009detection} used hand engineered features and supervised learning to detect abusive language. Other works~\citep{sood2012automatic} talk about why simply having a blacklist of words might not be enough to build a robust system to detect offensive content. There has been more recent works which focus on crowd sourcing annotations and hand engineering various features such as n-grams, syntactic and semantic features for this task~\citep{sood2012automatic,nobata2016abusive}. There has also been a line of work on extracting user level features to identify people exhibiting abusive behavior~\citep{chen2012detecting,buckels2014trolls,papegnies2017impact}. 

Twitter has been used by many researchers to detect abusive
language.~\cite{xiang2012detecting} created topical clusters over tweets with
a boot strapping technique.~\cite{davidson2017automated} performed an
in-depth analysis of hate speech on twitter, differentiating between hate speech
and offensive language. Recently there has been work on deep learning approaches
to identify abusive language.~\cite{badjatiya2017deep} used deep
learning based methods compromising of CNNs and LSTMs for hate speech detection
in Tweets. CNNs have also been used for hate speech classification~\citep{gamback2017using}. Additionally,~\cite{park2017one} proposed a 2 step
processes using hybrid CNNs.  One common issue is how to define abusive language
while annotating the data due to its subjectivity and lack of
context~\citep{ross2017measuring,schmidt2017survey}. ~\cite{waseem2016you}
showed that is harder for non-expert users to annotate such data reliably. We
ask the readers to refer to~\cite{schmidt2017survey} for a comprehensive survey
on this topic.
 
 We propose a data augmentation technique to generate a large corpus of data
 that can also capture nuances like ``recency'' and ``subjectivity'' for
 sensitive content. Since our approach is not model specific but complements the
 data used to train the sensitive content detection models, it can be combined
 with the aforementioned works for better performance.

\section{Data}\label{data}

\begin{figure}[]
\subfloat[{Size of Datasets visualized. Reddit is in the order of millions as compared to our other datasets which are in the order of thousands.}]{
  \includegraphics[width=65mm]{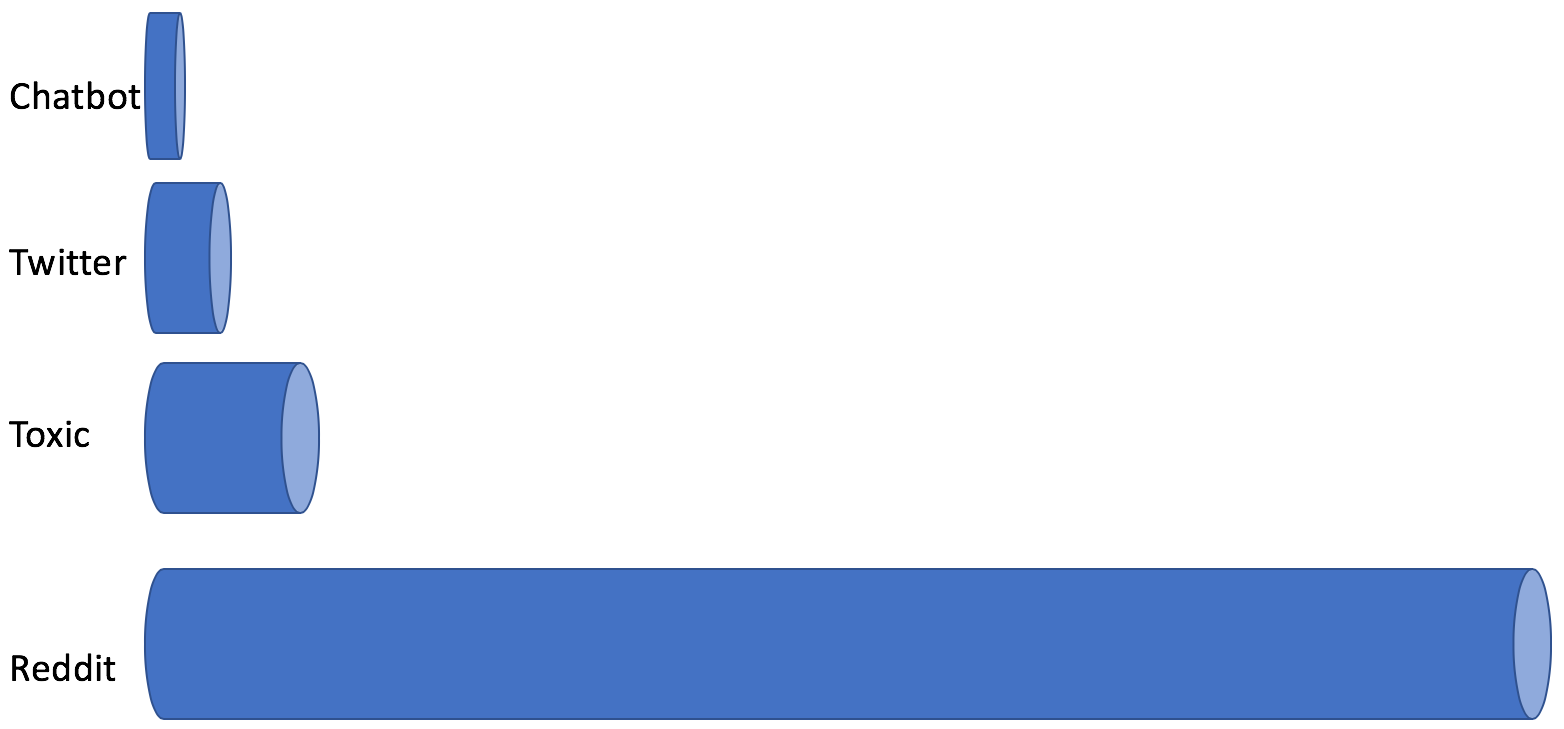}
  \label{fig:data_size}
  }
\subfloat[{All of our models are BiLSTM models.}]{
  \includegraphics[width=65mm]{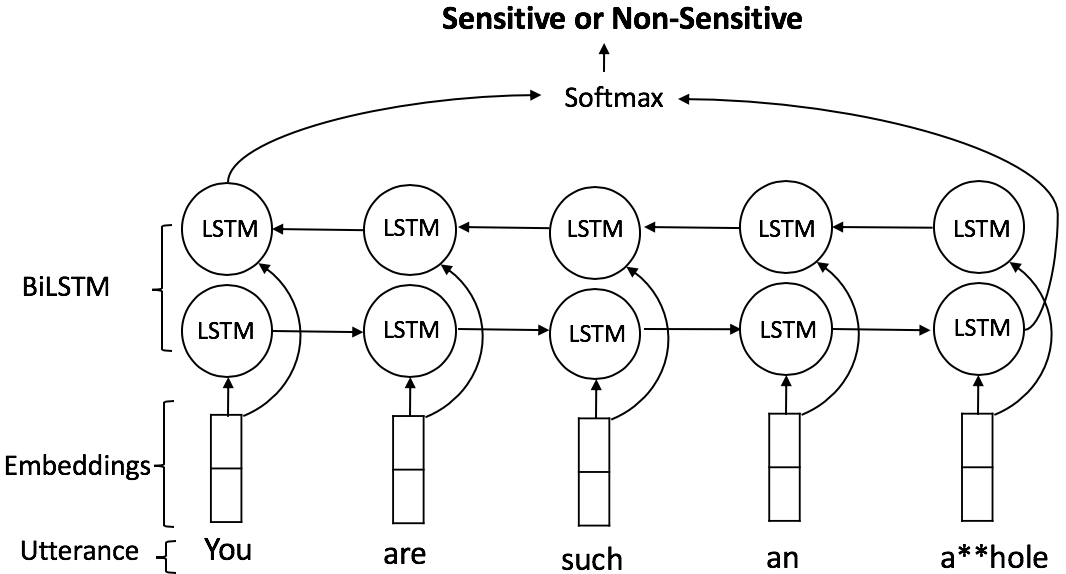}
  \label{fig:lstm}
  }
  \caption{}
\end{figure}

The two main types of datasets we use in this work are 1) small annotated datasets and 2) large unannotated datasets. Annotated datasets have challenges such as subjectivity and coverage. Additionally there is not a large corpus of well annotated data that fits the broad use case of identifying sensitive content. Therefore, we leverage our small annotated datasets to help select training data from our large datasets.  Additionally the level of sensitivity around a certain topic can rapidly change due to the Recency problem.
In our study we leverage a large amount of unannotated data and evaluate if it is useful due to the context certain utterances appear in. We test our models on in-domain and out-of-domain testsets where in-domain corresponds to training and evaluation on the same datasource while of out-of-domain corresponds to the model trained and evaluated on separate datasources. Our in-domain testsets consist of Twitter, Toxic, and Reddit. While our out-of-domain testset is an open domain dialog dataset which we will denote as Chatbot dataset. Figure~\ref{fig:data_size} visualizes the size of our datasets, to show the massive difference in size of Reddit as compared to our other datasets. A more detailed breakdown of the datasets can also be seen in Table~\ref{data_distribution}. We will now describe each of our datasets, including the data split and label space.

\begin{table}[!tbph]
  \caption{\label{data_dist} {Dataset Distribution.}}
  \centering
  \begin{tabular}{l|p{2.1cm}|p{1.8cm}|c|c|p{1.8cm}}
  \bf Data Source 
  & \bf \# of Sentences 
  & \bf \% Sensitive 
  & \bf \# Classes  
  & \bf Vocab Size
  & \bf Median Length (\#Words)
  \\ \hline
    Twitter & 24,783 &  83.0\% & 3 & 34,000 & 12 \\
    Toxic & 159,571 & 10.2\% & 7 & 70,198 & 23 \\
    Chatbot Dataset & 5,929 & 25.6 \% & 8 & 2,869& 4   \\
    Reddit & 20M  &  50.0\% & 2 & 1,192,334 & 11  \\
  \end{tabular}
  \label{data_distribution}
\end{table}

\subsection{Blacklist}\label{blacklist}
We started out with a blacklist of over 800 curated words. These words revolve around items such as profanity, hate, sexual content, and insults.
Because a blacklist is simply a list of tokens whose presence in a text implies offensive content, it can be considered both as a dataset and a model. Therefore we will also leverage the blacklist to help select utterances on large unannotated datasets as described in Section~\ref{blacklist-reddit}.

\subsection{Reddit Dataset}\label{reddit}
Reddit is a social news aggregation site that has many user discussion forums. These subreddits cover a variety of topics from news, movies, music and even offensive/sexual content.

Due to the online discussion forum format, a subreddit will have up to date context on a certain entity that annotations can not simply cover. 
In our work, we use Reddit as a static data dump\footnote{We use a publicly available Reddit data dump~\citep{reddit}.}, however this process can be converted to do real-time scraping of our training data to get even more up to date news.
Our dataset size contains 20M utterances categorized by over 5000 subreddits  where we had a 80\% split for train, 10\% split for development and 10\% split for test. Reddit is an unannotated dataset and the process to split the data into sensitive and non-sensitive classes will be discussed in Section~\ref{semi-supervised}.

\subsection{Toxic Comments Dataset}\label{toxic_data}
There has been work on trying to annotate datasets around sensitive content. The ``Toxic Comments Data'' challenge~\citep{toxic} was introduced on Kaggle, where the area of focus is negative online behavior. They deem these online behaviors as toxic utterances or utterances that are rude and disrespectful towards other users. About 150,000 Wikipedia utterances were shared, and were labeled in a multi-class format. The multiple classes were: toxic, severe toxic, obscene, threat, insult, identity hate and non-toxic. We took the labels: toxic, severe toxic, obscene, threat, insult, identity hate and grouped them together into a sensitive class. The non-toxic class became our non-sensitive class.

We will use this dataset to train a first-pass binary classifier, where the two classes are sensitive and non-sensitive.  We use a 90\% train split and 10\% test split. We will evaluate this model on its corresponding testset and on testsets that are out-of-domain with respect to this dataset. This first-pass classifier will also be used in our two stage semi-supervised data selection approach described in Section~\ref{weak-reddit}.

\subsection{Twitter Dataset}\label{public_data}
We trained on the public Twitter dataset \citep{davidson2017automated, twitter} which was collected and labeled using crowdsourcing.
This dataset contains three classes: Hate, Offensive, and Neither. We took the labels Hate and Offensive as our sensitive class and Neither was taken as our non-sensitive class. 
For Twitter, we build a binary classifier using a 90\% train split and 10\% test split. We will evaluate this model on its corresponding testset and on testsets that are out-of-domain with respect to this data set.

\subsection{Chatbot Test Dataset}\label{test_data}
We tested our model on utterances collected during a chatbot competition~\citep{ram2017alexaprize}. These utterances were manually annotated not just as sensitive or non-sensitive but had multi-class labels. The sensitive labels are: InappropriateIntent, InsultingIntent, Multiple InappropriateIntent, ProfaneIntent, RacistIntent, SexualIntent, and ViolentIntent. Examples of utterances are in Table~\ref{test_set_examples}. This test data was comprised of 4,412 utterances in the non-sensitive class and 1,517 utterances in the sensitive class. This was the only dataset was not used to train a model and is only used to test our sensitive classifier. 

\begin{table}[!tbph]
  \caption{\label{test_set_examples} {Examples of Sensitive Content in Chatbot test set}}
  \centering
  \begin{tabular}{c|c}
  \bf Chatbot Labels & \bf Chatbot Sensitive Examples \\ \hline
    InappropriateIntent & tell me something dirty\\
    Multiple InappropriateIntent & uhhh my my life goal is to get rip this f*** and my hobby is j****** off \\
    InsultingIntent & you're stupid \\
    ViolentIntent & i will m***** you \\
  \end{tabular}
\end{table}

\section{Two Stage Semi-Supervision}\label{semi-supervised}
Due to the absence of large scale annotated data, we created a large set of training data using Reddit in a two stage approach. As described in Section~\ref{reddit} Reddit is comprised of many different subreddits. A subreddit is a user-created discussion forum that typically covers a certain topic which ranges from news, movies, music and to offensive/sexual content.

\begin{figure}[t]
\caption{\label{fig:twostage} {\textbf{Two-stage Semi-Supervised Pipeline.} Stage 1 consists of sorting subreddits with our blacklist, by simply counting the number of words in said blacklist. During Stage 2, high confidence sensitive comments were sampled either if the probability from the toxic classifier was greater than 0.8 or if they contained words from the blacklist. For low confidence sensitive, comments were sampled if the probability from the toxic classifier was less than 0.3, and making sure no words were in the blacklist.}}
\centering
\includegraphics[height=45mm]{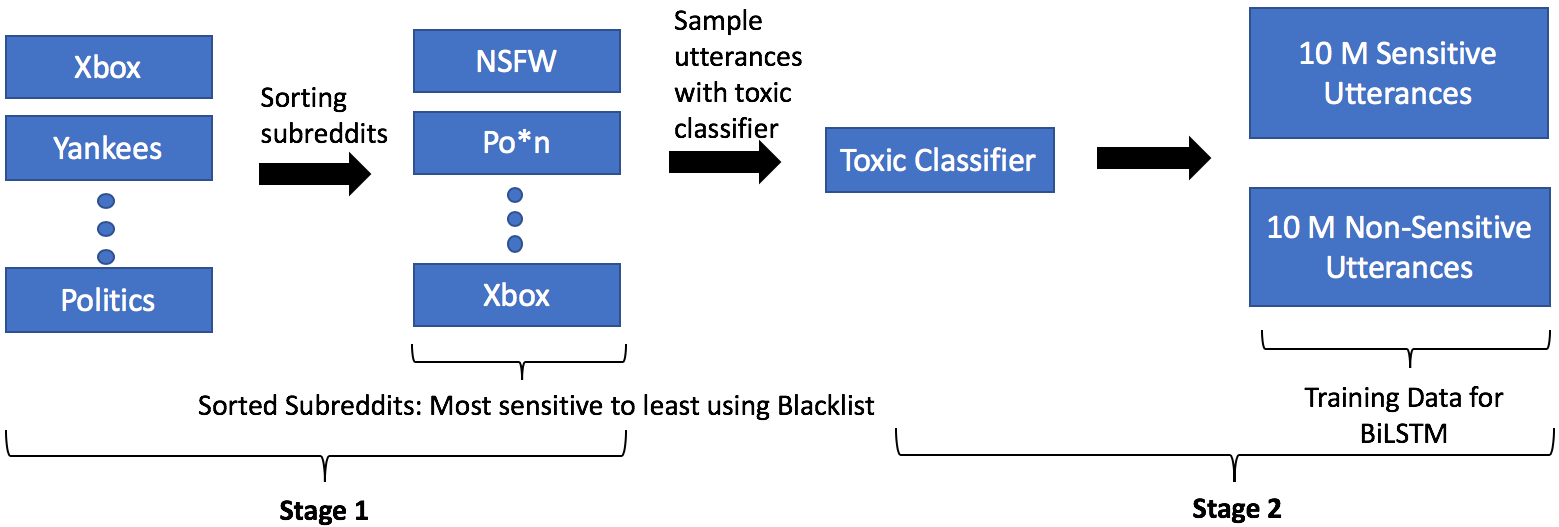}
\end{figure}

 Stage 1 utilizes our blacklist to sort our subreddits from most sensitive to least sensitive. We will then sample utterances from Stage 1 for our first model. Stage 2 utilizes both our blacklist and a weak supervised model to select utterances from these subreddits for our second model. The entire pipeline can be seen in Figure~\ref{fig:twostage}.

\begin{table}[!tbph]
  \caption{\label{subreddits} {Subreddits sorted by percentage of blacklist words. The score indicates percentage of blacklist words in the subreddit divided by total number of words in said subreddit.}}
  \centering
  \begin{tabular}{c | c}
  \bf Most Sensitive Subreddits & \bf Least Sensitive Subreddits\\ \hline
     9.70\% &  0.2\% \\
    5.53\% &  0.09\%\\
    4.78\% &  0.11\%  \\
  \end{tabular}
\end{table}

\subsection{Stage 1: Blacklist based Reddit sampling}\label{blacklist-reddit}

For over 5000 subreddits in our dataset, we obtained the percentage of sensitive words in our subreddits. This percentage was calculated as the number of words in the subreddit being in the blacklist divided by the total number of words in the subreddit. We then sort on this percentage in descending order. A table of these percentages can be seen in Table~\ref{subreddits}. The assumption here is that subreddits that have a high percentage of words from the blacklist, have other comments within them that are potentially sensitive.

We will then sample from the most sensitive subreddits and the most non-sensitive subreddits to create our training set. For our non-sensitive class, we sampled from 2,802 subreddits that had a percentage < 0.2\%. For our sensitive class, we sampled from 144 subreddits that had a percentage > 1.0\%. The model which we trained with this dataset will be denoted as Bootstrap. For Bootstrap, we build a binary classifier using a 80\% train, 10\% dev and 10\% test split. 

\subsection{Stage 2: Weakly Supervised Model based Reddit sampling}\label{weak-reddit}
Using the Toxic Comments Data as described in Section~\ref{toxic_data}, we train a binary classifier. Table 4 shows the performance of this classifier. Due to its good performance, we then used this classifier to sample utterances from our sorted subreddits list as described in Section~\ref{blacklist-reddit}. We sample 10 Million high confidence Reddit utterances from the most sensitive subreddits. We sample 7 million utterances that gave a probability greater than 0.8 from the toxic classifier and 3 million utterances which contain words from the blacklist. We also sample 10 Million high confidence non-sensitive utterances from the least sensitive subreddits. We only took utterances which gave a probability of less than 0.3 from the toxic classifier and contained no words in the blacklist. The total dataset was 20 Million utterances.
The model which we trained with this dataset will be denoted as Two-Stage(TS) Bootstrap or TS Bootstrap for short. For TS Bootstrap, we build a binary classifier using a 80\% train, 10\% dev and 10\% test split.

\section{Experimental Setup}\label{models}

All of our models are BiLSTM models as shown in Figure~\ref{fig:lstm}. More formally, assume an input utterance of length $L$ and corresponding $D$ dimensional word embeddings $e_i, i=1, \cdots L$, then the network structure is: 
\begin{gather*}
h_{f}, h_{b}  = BiLSTM({e_i}) \\
Sensitive/NonSensitive = softmax({[h_{f}; h_{b}]})
\end{gather*}
$h_{f}$ and $h_{b}$ correspond to the final states of the forward and backward LSTM respectively and after concatenation are sent through a softmax layer for binary classification.
 
Each dataset except for Chatbot in Section~\ref{data} will be used as training data for our models. Each model will then be tested against the other datasets to see generalizability on out-of-domain datasets.  All of our models are BiLSTM models which used a word embedding size of 300. Word embeddings were initialized with Glove embeddings~\citep{glove2014}. We used the Adam optimizer with a weight decay of 1e-8, and other parameters are defaults. We use a dropout rate of 0.5, a learning rate of 0.001 and a batch size of 500.

\section{Results and Discussion}\label{results}

We perform a variety of experiments on in-domain and out-of-domain scenarios, where in-domain corresponds to training and evaluation on the same datasource while of out-of-domain corresponds to the model trained and evaluated on a separate datasources.  For all our tables, the training dataset column refers to the training data for the model. Bootstrap and TS Bootstrap correspond to the two stage semi-supervised approach as described in Section~\ref{semi-supervised}. Also combined training data from various datasets is denoted with a ``+'' such as ``+ T + TX'' in the training dataset column. Table~\ref{all_results} depicts the performance of all the in-domain and out-of-domain settings. We see that even though the bootstrap datasets are out-of-domain for Toxic and Twitter testsets, they significantly out-perform the models trained on in-domain data. Performance of the bootstrap models does improve slightly by adding Toxic and Twitter training datasets on Toxic testset, however the performance degrades slightly in the case of Twitter testset. Bootstrap models outperform Twitter and Toxic models by large margins on the Chatbot testset. One of the possible reasons is Reddit data is both conversational and topical similar to our Chatbot testset. The Chatbot testset is not affected much by adding Twitter and Toxic datasets, since the Chatbot testset is out-of-domain for all the settings. Furthermore, performance of the Bootstrap model alone improves significantly compared to the Two Stage Bootstrap model by adding Toxic and Twitter training datasets, which may be because of the fact that training data for Two Stage Bootstrap is sampled with Toxic Model based classifier. Through these observations we note that the bootstrap models generalize well across various domains.

\begin{table}[h]
\caption{\label{all_results} {F1-scores and Accuracy scores across various testsets.}} 
\begin{center}
\begin{tabular}{l | c | c | c | c | c | c }
Training Dataset
& \multicolumn{2}{c|}{Twitter Testset}
& \multicolumn{2}{c|}{Toxic Testset }
& \multicolumn{2}{c}{Chatbot Testset}                \\\hline
       & F1-score & Accuracy & F1-score & Accuracy & F1-score & Accuracy  \\\hline
Twitter: T & 87.1 & 86.5 & 85.4 & 82.6 & 74.9 & 73.9  \\
Toxic: TX & 82.8 & 81.8 & 90.2 & 90.5 & 74.0 & 85.0  \\
Bootstrap & 84.0 & 82.1 & 91.4 & 92.0 & 92.0 & 90.0  \\
Bootstrap + T + TX & 87.8 & 86.8 & 95.4 & \textbf{95.7} & 92.1 & 92.0 \\
TS Bootstrap & \textbf{88.7} & \textbf{87.7} & 94.4 & 94.0 & 95.5 & \textbf{96.0} \\
TS Bootstrap + T + TX & 88.3 & 87.6 & \textbf{95.5} & 95.6 & \textbf{95.6} & 95.6 \\
\end{tabular}
\end{center}
\end{table}

A breakdown of our metrics for our Chatbot testset are shown in Table~\ref{sensitive_class_chatbot}. The baseline models such as the blacklist have high precision on the sensitive class but have poor recall on the same class. These are utterances that are not being captured because they do not have obvious profane or offensive language leading to poor F1-scores. The Bootstrap model improves upon the recall significantly. We observe that toxic based model also performs better than the blacklist, which led us to leverage this model for further data selection on our Reddit dataset. Finally the Bootstrap TS model improves recall of the sensitive class significantly leading to an F1-score of 95.5. Combining the training data from TS Bootstrap with Twitter and Toxic improves our F1-score to 95.6. 

\begin{table}[h]
\caption{\label{sensitive_class_chatbot} {Sensitive Classifier Results on Chatbot testset. We measure Precision and Recall for our Sensitive class(S) and Non-Sensitive class(NS)}} 
\begin{center}
\begin{tabular}{ l | c | c | c | c  }
Training Dataset 
& Precision (S)
& Precision (NS)
& Recall (S)
& Recall (NS)
\\\hline
Blacklist & \textbf{99.4} & 79.4 & 24.6 & \textbf{99.9} \\
Twitter: T & 49.2 & 86.5 & 65.3 & 76.8 \\
 Toxic: TX  & \textbf{99.5} & 82.0 & 41.0 & \textbf{99.9} \\
 Bootstrap & 85.4  & 94.3 & 85.4 & 95.0  \\
 Bootstrap + T + TX & 85.7 & 94.2 & 82.9 & 95.2  \\
 TS Bootstrap & 90.1 & \textbf{97.2} &  \textbf{92.0} & 97.0 \\
 TS Bootstrap + T + TX & 91.5 & 97.0 & 91.3 & 97.1  \\
\end{tabular}
\end{center}
\end{table}

In Table~\ref{avg_recall}, we show weighted average recall results across various testsets. Recall is a very important metric for this task, as it refers being able to capture sensitive utterances that do not have obvious features in them that make them sensitive. Twitter, Toxic and Chatbot datasets are annotated with multiple classes of sensitivity such as profanity, hate speech, sexual content, racist, etc. We evaluated all the models on weighted average recall across all the classes. We show that our methods (trained on out-of-domain data) perform significantly better than the models trained on in-domain datasets. That is large scale two stage semi-supervision can help recover sensitive content from a variety of classes without explicitly providing the labels for these classes.

\begin{table}[h]
\caption{\label{avg_recall} {Weighted Average Recall across various testsets.}} 
\begin{center}
\begin{tabular}{ l | c | c | c  }
Training Dataset
& {Twitter Testset}
& {Toxic Testset}
& {Chatbot Testset} \\\hline
Twitter: T & 84.5 & 82.6 & 73.2 \\
Toxic: TX & 81.8 & 85.5 & 92.7\\
Bootstrap & 82.1 & 92.0 &  90.2  \\
Bootstrap + T + TX & 86.9 & 81.7 & 91.9\\
TS Bootstrap & \textbf{89.5} & \textbf{94.0} &  94.5  \\
TS Bootstrap + T + TX & 87.8 & 88.6 & \textbf{95.5} 
\end{tabular}
\end{center}
\end{table}

\begin{table}[!tbph]
\caption{\label{examples} {Qualitative Examples to showcase the power in leveraging a very large dataset that can take into account context to help capture sensitive utterances.}} 
\begin{center}
\begin{tabular}{l c c c c c }
Sensitive Utterances
& Blacklist
& {Twitter}
& {Toxic}
& {Bootstrap}
& {TS Bootstrap} \\\hline
\begin{tabular}[c]{@{}l@{}}``i **** somethin' out this \\ mornin prettier than you''\end{tabular} & \checkmark  & \checkmark & \checkmark & \checkmark & \checkmark \\
``They are *** and ****'' & x  & \checkmark & \checkmark & \checkmark  & \checkmark \\
``Z is the devil'' & x  & x & x & \checkmark  & \checkmark \\
``X is the most elaborate lie ever told'' & x  & x & x & x & \checkmark \\
``mate, are you Y? ******* is in the air'' & x  & x & x & x & \checkmark \\
``why don't you get a life you sicko'' & x  & x & x & x & \checkmark \\
\begin{tabular}[c]{@{}l@{}}"You've done so well for someone \\ with your education level.''\end{tabular} & x  & x & x & x & x \\
\end{tabular}
\end{center}
\end{table}

We also show some qualitative results of our models in Table~\ref{examples}. It can be observed that large scale semi-supervised techniques are able to identify really hard cases such as "Z is the most elaborate lie ever told" and "X is the capital of Y"\footnote{We have removed actual words for sensitivity}. Some of these sensitive cases are actually incorrect facts. Since the models are trained on large scale conversational data which is topical, the models appear to have learned that the combination of X, Y and capital are associated with a significantly higher probability of sensitive content. Identifying cases such as this presents a significant challenge even for human annotators, given the subtle cultural context involved. We also show an example that could not be caught by any models, because it requires the model to understand sarcasm which can be used to hide the abusive statement.

\section{Conclusion and Future Work}
Sensitive content detection is a nuanced problem which also suffers from lack of standard definitions and datasets. Through this work we provide a simple technique to bootstrap large amounts of semi-supervised data, leading to very good performance on a variety datasets. We show that our data selection strategy leads to significant performance boost in both in-domain and out-of-domain settings. We show that the blacklist based approach doesn't scale well. Adding more blacklist words (such as "devil") leads to false positives while not adding them leads to poor recall. Since our data captures the implicit details about sensitive speech, we observe that we are able to achieve very high recall on the sensitive class and capture subtle political messages which makes a sentence sensitive. Furthermore, we also showcase that models trained using large scale semi-supervised data have high recall on varieties of sensitive classes such as hate speech, racism, sex and pornography, etc. without providing explicit labels for these classes. Our proposed data selection strategy can be combined with different modeling techniques (such as the ones mentioned in Section~\ref{related}) to build more powerful classifiers which we plan to do in the future.

\small
\bibliographystyle{acl_natbib.bst}
\bibliography{refs.bib}

\end{document}